%% file: 0_main_Autoscaling.tex
\documentclass[letterpaper]{article} 
\usepackage[]{aaai23}  
\usepackage{times}  
\usepackage{helvet}  
\usepackage{courier}  
\usepackage[hyphens]{url}  
\usepackage{graphicx} 
\usepackage{amsmath}
\usepackage{natbib}  
\usepackage{caption} 
\usepackage{algorithm}
\usepackage{algorithmic}
\usepackage{subfigure}

\usepackage{newfloat}
\usepackage{listings}
\DeclareCaptionStyle{ruled}{labelfont=normalfont,labelsep=colon,strut=off} 
\floatstyle{ruled}
\newfloat{listing}{tb}{lst}{}
\floatname{listing}{Listing}
\title{AHPA: Adaptive Horizontal Pod Autoscaling Systems on Alibaba Cloud Container Service for Kubernetes}

\author{
    Zhiqiang Zhou\textsuperscript{\rm 1},
    Chaoli Zhang\textsuperscript{\rm 1},
    Lingna Ma\textsuperscript{\rm 1},
    Jing Gu\textsuperscript{\rm 3},
    Huajie Qian\textsuperscript{\rm 2},
    Qingsong Wen\textsuperscript{\rm 2},\\
    Liang Sun\textsuperscript{\rm 2},
    Peng Li\textsuperscript{\rm 3},
    Zhimin Tang\textsuperscript{\rm 3}
}
\affiliations{
    \textsuperscript{\rm 1}DAMO Academy, Alibaba Group, Hangzhou, China\\
    \textsuperscript{\rm 2}DAMO Academy, Alibaba Group, Bellevue, WA, USA\\
    \textsuperscript{\rm 3}Alibaba Cloud, Alibaba Group, Hangzhou, China\\

\{zhouzhiqiang.zzq, chaoli.zcl, malingna.mln, zibai.gj, h.qian, qingsong.wen, liang.sun, yuanyi.lp, zhimin.tangzm\}@alibaba-inc.com
}

\begin{document}

\maketitle

\begin{abstract}
The existing resource allocation policy for application instances in Kubernetes cannot dynamically adjust according to the requirement of business, which would cause an enormous waste of resources during fluctuations. Moreover, the emergence of new cloud services puts higher resource management requirements. This paper discusses horizontal POD resources management in Alibaba Cloud Container Services with a newly deployed AI algorithm framework named AHPA - the adaptive horizontal pod auto-scaling system. Based on a robust decomposition forecasting algorithm and performance training model, AHPA offers an optimal pod number adjustment plan that could reduce POD resources and maintain business stability. Since being deployed in April 2021, this system has expanded to multiple customer scenarios, including logistics, social networks, AI audio and video, e-commerce, etc. Compared with the previous algorithms, AHPA solves the elastic lag problem, 
increasing CPU usage by 10\% and reducing resource cost by more than 20\%. In addition, AHPA can automatically perform flexible planning according to the predicted business volume without manual intervention, significantly saving operation and maintenance costs.
\end{abstract}

\input{1_introduction}

\input{2_application_description}

\input{3_useofAI_technology}

\input{4_experimental_evaluation}

\input{5_application_use_payoff}

\input{6_application_development_deployment}

\input{7_maintenance}

\input{8_conclusion}



\bibliography{aaai23}

\end{document}

%% file: 1_introduction.tex
\section{Introduction}





The continuous development of cloud computing technology provides more possibilities for current computer online services, and users also have higher expectations for cloud resilience. Furthermore, with the development of container services in cloud-native technologies~\cite{brewer2015kubernetes,balalaie2016microservices}, more and more new types of applications based on container services have emerged. 
Unlike applications in the Virtual Machine~\cite{altintas2005virtual} era, where minute-level manual operation is enough, new emerging applications usually require second-level operations. 
Meanwhile, 
many fast-developing applications have noticeable cyclical fluctuations~\cite{yan2021workload,higginson2020database,cortez2017resource,atikoglu2012workload,SPAR08}, 
such as Internet broadcast, e-Learning, online game, etc. 
This kind of application repeatedly emerges with peaks and valleys in business demand, so flexible resource utilization with low latency is required. 
Another main kind of new application is serverless computing~\cite{baldini2017serverless,mcgrath2017serverless} which allows clients to build and run services without thinking about servers. It is challenging to elastically manage resources for such applications facing the difficulty of dealing with cold starts, low latency, and scaling efficiency. Thus, proactive autoscaling is popular~\cite{taft2018pstore,rebjock2020simple,rzadca2020autopilot,tsoumakos2013automated}.

Both cloud technological maturity and the emergence of new businesses have led to the development of efficient resource utilization. More specifically, new application characteristics call for a unique design of auto-scaling, which allows adjustment of application instances to improve the utilization under the cloud native concept.
Kubernetes (k8s)~\cite{burns2022kubernetes} is the primary open-source container orchestration system for application/software deployment and management maintained in the cloud native computing foundation. Scalability is one of the core requirements of Kubernetes (k8s). 
Auto-scaling~\cite{lorido2014review,qu2018auto} is a necessary feature of the Kubernetes platform to secure scalability. Auto-scaling saves time, prevents performance bottlenecks, and avoids resource waste with appropriate configurations. 

\begin{figure}[t] 
\centering 
\includegraphics[width=0.48\textwidth]{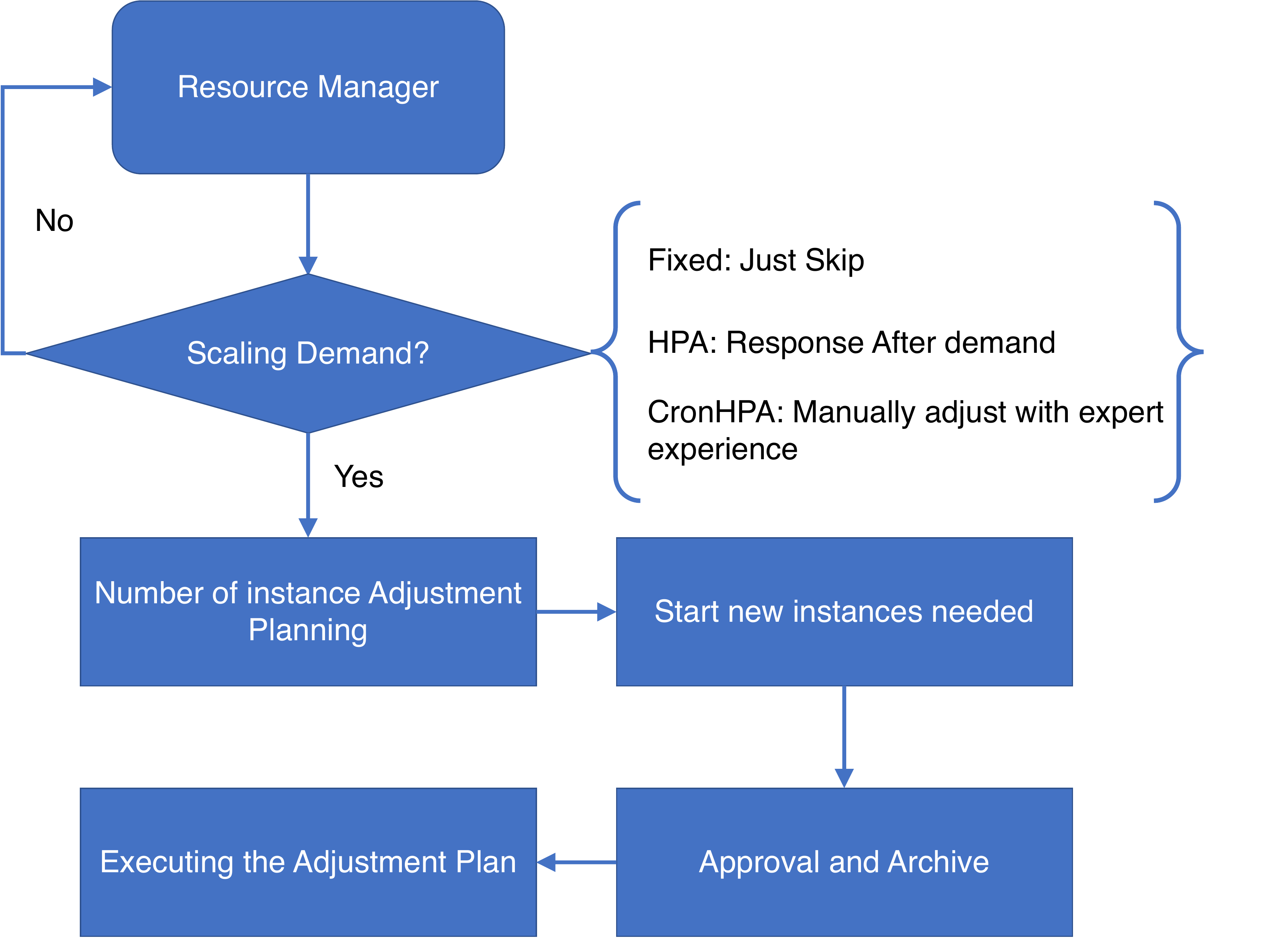} 
\caption{The procedure of resource management in Kubernetes with three conservative ways (Fixed, HPA, and CronHPA).} 
\label{conservative process} 
\end{figure}

Currently, there are three conservative ways to manage the number of application instances in Kubernetes, whose process is shown in Fig.~\ref{conservative process}: fixed number of instances, HPA~\cite{nguyen2020horizontal}, and CronHPA\footnote{https://www.alibabacloud.com/help/en/container-service-for-kubernetes/latest/cronhpa-186978}. Unlike the fixed number of instances which does not change when fluctuations of demand occur, HPA and CronHPA can adjust the number of instances according to the change in demand. However, they have shortcomings in efficiency, simplicity, and accuracy, and none of these methods could resolve the demand fluctuations elastically. 
For further details, the policy with a fixed number of instances is the easiest to implement and widely used. Meanwhile, it also has a significant disadvantage of wasting resources in the valleys of business demand. Compared with the static method, the HPA method adjusts the number of instances after the demand varies. Hence, its response to fluctuations in demand is lagging, which may lead to ineffective processing and lousy quality of service, and even worse, the application is terminated. In addition, the CronHPA policy requires expert experience to manually set up the scaling schedule, which might be inaccurate, inflexible, and a massive cost to human resources.


To address the shortcomings of existing solutions mentioned above,  
we designed and deployed a new system named the Adaptive Horizontal Pod Auto-scaling System(AHPA) in the product of Alibaba Cloud Container Service for Kubernetes (ACK\footnote{https://www.alibabacloud.com/product/kubernetes}) that supports much better predictive auto-scaling. AHPA solves the problem of existing methods not being able to dynamically adjust POD resources by using a decomposition method to accurately predict the next phase of business. At the same time, AHPA learns the mapping from business workload to the number of required PODs, 
and provides the final scaling action plan, unlike CronHPA which requires manual intervention. The automated model makes AHPA easy to deploy and scale up, and saves significant operation and maintenance costs.


AHPA system has been deployed throughout Alibaba Cloud Service since April 2021. It has significantly improved the elastic resource management compared to the previous algorithms used in Cloud Service. More specifically, in the scenario of intelligent voice, AHPA's current call reaches more than 5000 times per day. It has saved POD resources cost about ten thousand daily, directly contributing to more than 28\% daily cost savings compared to the original.

%% file: 2_application_description.tex
\section{Application Description}

\begin{figure}[t] 
\centering 
\includegraphics[width=0.43\textwidth]{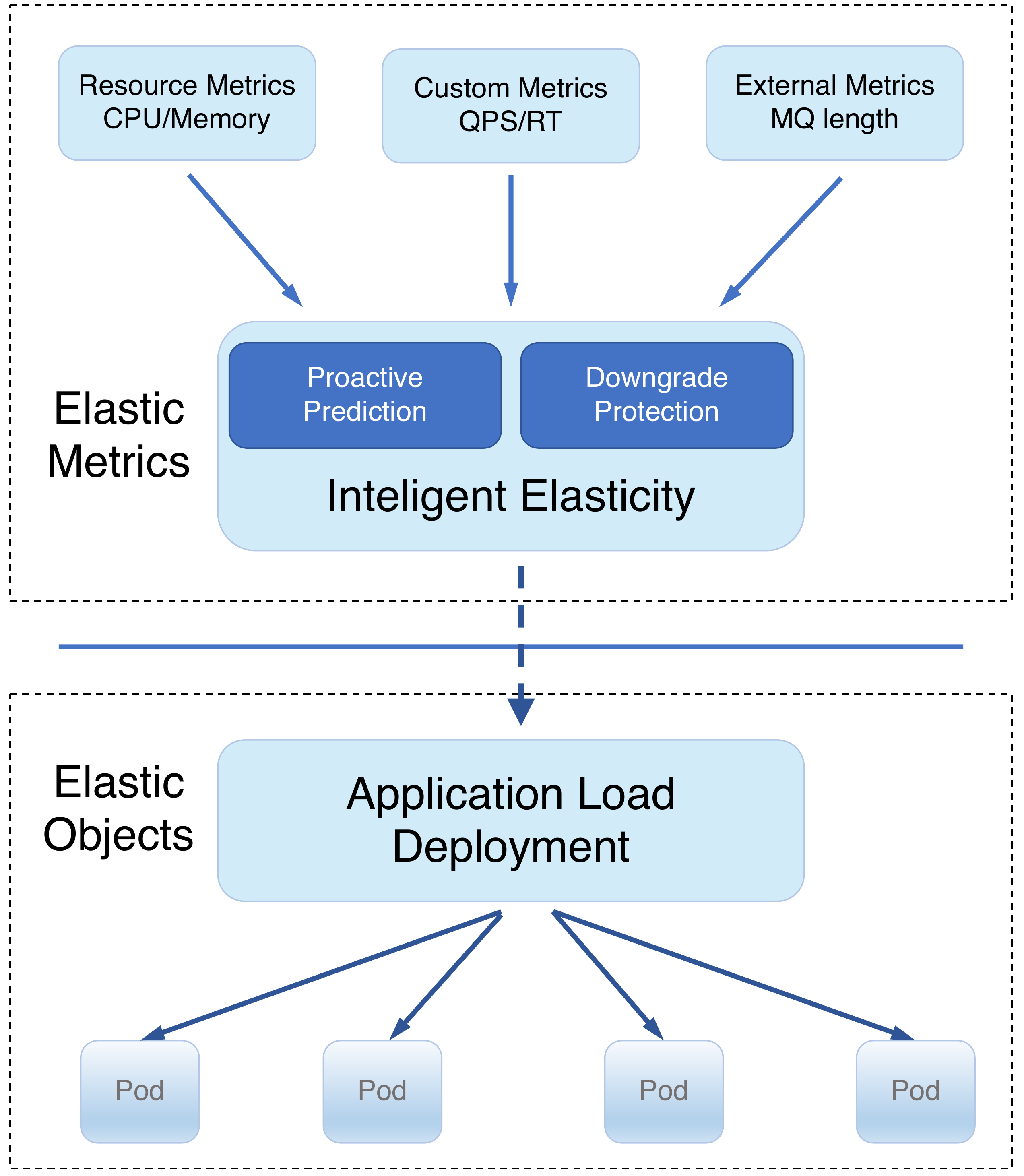} 
\caption{The architecture of the AHPA system.} 
\label{Art} 
\end{figure}

In this section, we introduce the architecture of the AHPA system, as illustrated in Fig.~\ref{Art}. It consists of two main parts: elastic metrics and elastic objects. 

In elastic metrics, various metrics are provided to satisfy different demands of clients, including resource metrics (for instance, CPU and memory), custom metrics (for instance, query rate per second and response time), and external metrics (for instance, message queue length). 
A proactive prediction module is used to predict demand in the future, and a downgrade protection module is designed to guarantee stability, hence intelligent elasticity module outputs resource prediction. With prediction results, the elastic objects module would allocate the required number of pods via application deployment. 

The primary purpose of elasticity resource management is to save costs for the client when satisfying the stability requirement of service and reduce the cost of human resources in operation and maintenance. 
Designed for the requirements mentioned above, our new algorithm AHPA has three typical features so it could greatly meet both the elastic demand and the stable requirement: 
\begin{itemize}
    \item Stability: AHPA is performed under the condition that the stability of client services is guaranteed.
    \item Zero cost in operation and maintenance: no additional operation or maintenance is needed, including no more added controllers and more concise configurations than HPA. 
    \item Serverless application feasibility: AHPA provides resources at the pod level without considering the usage rate at the node level, which enhances the long run of applications. 
\end{itemize}

%% file: 3_useofAI_technology.tex
\section{Use of AI Technology}
\begin{figure}[t] 
\centering 
\includegraphics[width=0.46\textwidth]{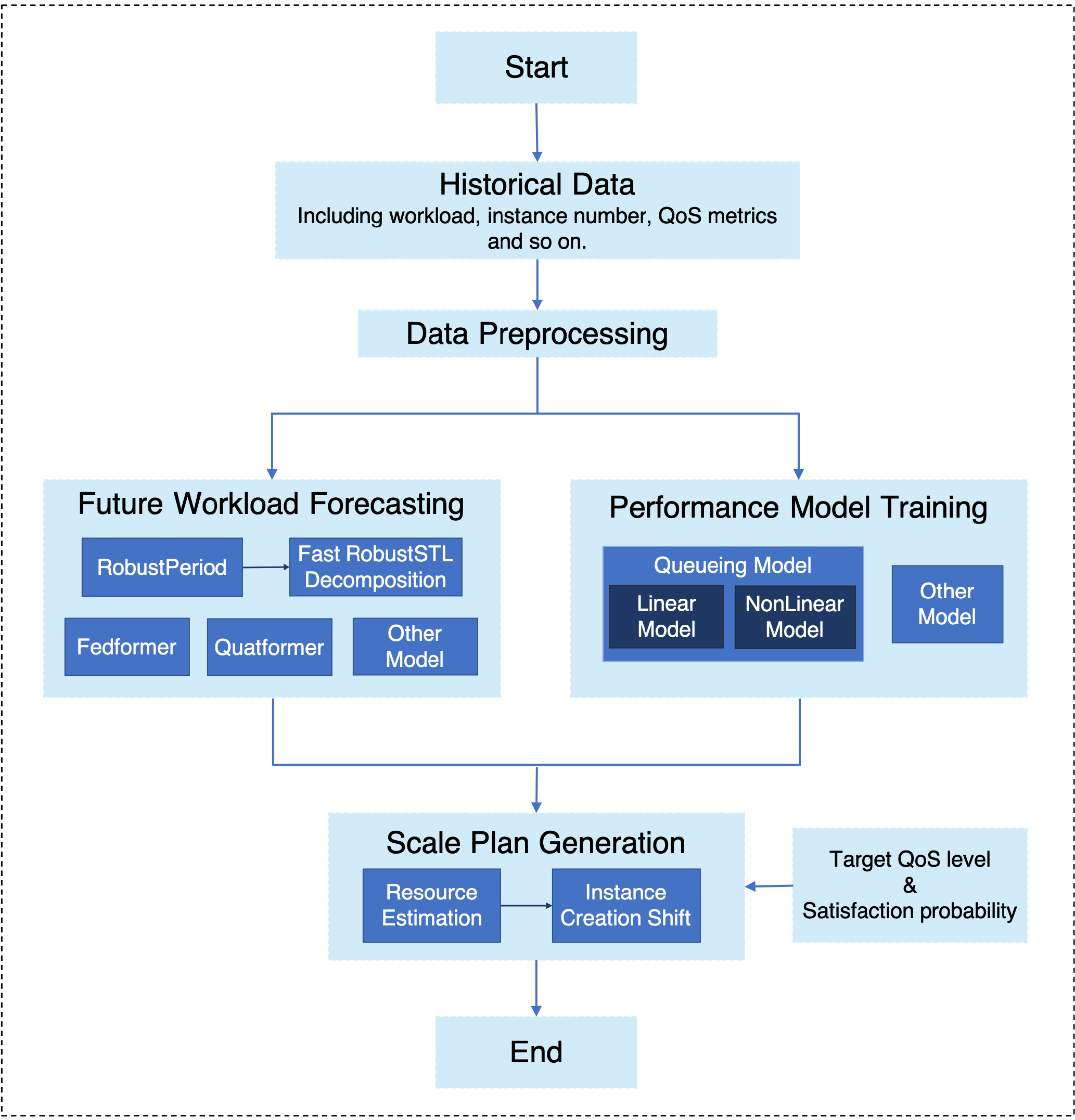} 
\caption{The framework of the scaling algorithm in AHPA.} 
\label{Alg} 
\end{figure}

The use of AI technology mainly focused on the following three modules in the adaptive horizontal pod auto-scaling framework as illustrated in Fig.~\ref{Alg}: future workload forecasting, performance model training, and scaling plan generation. In the following, we introduce each component in detail.

\subsection{Future Workload Forecasting}
The first module, named future workload forecasting, heavily relies on AI technology and plays an essential role in the AHPA framework. Accurate prediction of future workload would significantly improve the optimality of the schedule of the horizontal pod plan and hence earn more profit from cost savings in the system.
However, the forecasting process still faces several complex challenges in the Alibaba Cloud Service:
\begin{itemize}
    \item Missing values and noisy data: there are many possible causes for this issue. For example, when some nodes of the cloud service distributed system is damaged, or some accidents happen during the interaction between the user and the service system, they will lead to the lack of valid data. Besides, if the collected data is too far from the current time, it may affect forecasting accuracy. Therefore, setting the threshold to judge data validity is a necessary procedure. There is also a need for an appropriate method to complement the missing data and normalize the data scale in some situations.
    \item Limited data scale: metrics data in Kubernetes generally uses Prometheus storage. Considering cost and efficiency in a compromise, the general business data storage period is 7 days. The 7-day data volume is too small as a training set, and the trained machine/deep learning model usually has a poor accuracy. How to effectively estimate the future business volume with a limited amount of historical data is worthy of discussion.
    \item High complexity: in general, user demands change frequently, which significantly enriches the complexity of data. For instance, data may have complicated characteristics such as multiple cycles. Therefore, sophisticated data has higher requirements on the ability of the algorithm model to make an accurate prediction.
\end{itemize}

\begin{figure*}[t] 
\centering 
\includegraphics[width=0.85\textwidth,height = 6cm]{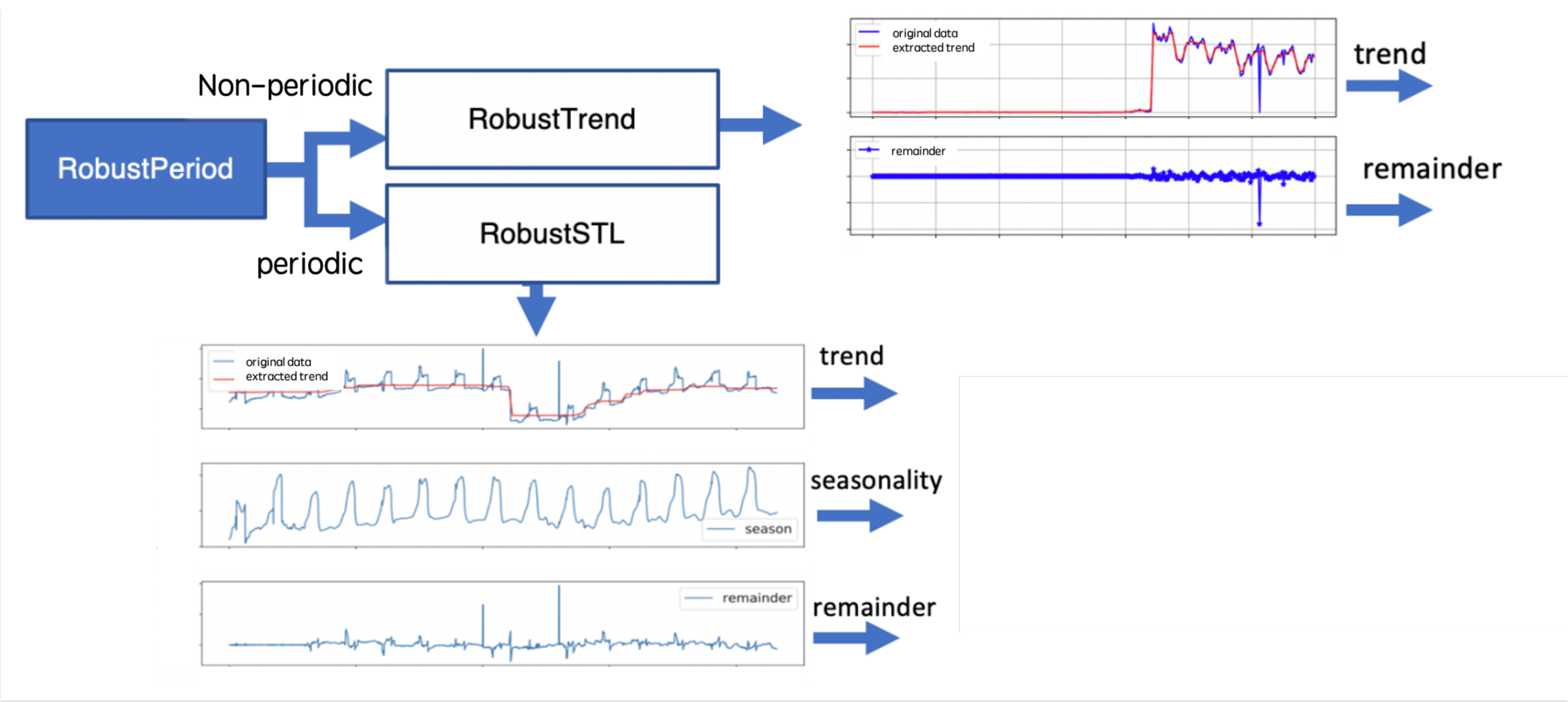} 
\caption{The robust time series decomposition for forecasting module in AHPA.} 
\label{forecasting framework} 
\end{figure*}

We design a robust decomposition-based statistical method as the main forecasting scheme to address the above challenges and cater to high forecast latency requirements. Specifically, we adopt our previously published three robust time series decomposition algorithms~\cite{WenRobustPeriod20,wen2019robuststl,wen2019robusttrend} as the preprocessing steps as shown in Fig.~\ref{forecasting framework}. Firstly, RobustPeriod~\cite{WenRobustPeriod20}, based on MODWT (maximal overlap discrete wavelet transform) to decouple multiple periodicities, is utilized to detect if the input time series is periodic and its period lengths.
According to the period detecting result, there are two ways to deal with different data characteristics. For periodic data, RobustSTL~\cite{wen2019robuststl,wen2020fast} is adopted to decompose 
the input time series into trend, seasonality (periodic component), and residual terms. For non-periodic data, RobustTrend~\cite{wen2019robusttrend} is adopted to decompose 
the input time series into trend and residual terms.






Mathematically, the above robust forecasting module decomposes the time series data into the trend item $\tau_t$, the period item $s_{i,t}$ (if it is periodic), and the residual item $r_t$, and the formula is as follows:
\begin{equation}
    y_t=\tau_t+\sum_{i=1}^ms_{i,t}+r_t,t=0,1,...,N-1
\end{equation}
The historical period item $s_{i,t}$ is directly shifted to the right as the prediction of the future period item. The trend term $\tau_t$ is predicted by a classical time series model such as exponential smoothing to obtain the prediction of the future trend component. Finally, the residual part uses the quantile regression forest to get an upper bound prediction of the future residuals. The combination of the above three items leads to the final predicted value $y_{t+1}$ at the next moment:
\begin{equation}
\begin{split}
    y_{t+1}=\sum_{i=1}^ms_{i,t}+\text{ExponentialSmoothing}(\tau_t)\\
    + \text{QuantileRegression}(r_t)
\end{split}
\end{equation}

In addition to the robust decomposition-based forecasting schemes introduced above, we also consider the state-of-the-art transformer-based deep learning forecasting models~\cite{wen2022transformers} for scenarios with enough data. Specifically, we include our recently developed FEDformer~\cite{zhou2022fedformer} model, which is suitable for long-term time series prediction scenarios, and our recently developed Quatformer~\cite{chen2022quartformer} model, which is mainly designed for data characteristics with complicated periodical patterns. Shortly we would continue working on the whole scheme and add more forecasting methods that could fit different situations.


\subsection{Performance Model Training}
After getting the forecast volume values, the second module, called the Performance Model, is used to simulate the relationship between the indicator metric and the number of pods. In this section, due to the reality of some specific businesses, we have assumed that pods are equivalent computing units. The model mainly adopts the method of queuing theory in operation research, including two kinds of different queuing models. One is the linear relation using parallel M/M/1 queues, and the other is M/M/c queues with a public buffer pod. 


In the Pod resource utilization queuing model, we regard the business QPS (Queries Per Second)  as the queue rate in the system and the number of Pods as the number of service desks $c$ (in the M/M/1 queues, each pod could be considered as a private desk and they have the same processing rate), and $u_i$ is the average service of the service desks which can be expressed in terms of Pod CPU. Based on these metrics, we aim to find the average wait time per customer consistent with the RT per request. Different average RT with different queue models could be formalized by the following mathematical formula:
\begin{equation}
\begin{split}
    \text{Avg} RT(M/M/1) = (u - \dfrac{QPS}{N})^{-1}+\text{otherlatency}\\
    \text{Avg} RT(M/M/c) = f(QPS,u,N)+\text{otherlatency}
\end{split}
\end{equation}
where $u$ denotes the queries per second that could be processed by one pod, and the $f$ is given by Erlang's C formula~\cite{2011A} and Little's Law~\cite{2008Little}. In general, we should find the minimum value of the pod number while satisfying the requirement of average RT. For the linear M/M/1 model, $pod = C* QPS$; and for the nonlinear M/M/c queue model, $pod = g(QPS)$. When the number of adjustable pods is small, the M/M/c model performs better than the other model, and while there are more pods available, M/M/1 queue model outperforms. Therefore, different models could be selected according to different requirements in actual deployments. In summary, this performance training module takes the predicted values of the previous forecasting module and historical indicator data as input and outputs the number of pods that need to be adjusted.

\subsection{Scaling Plan Generation}

Our system generates the final scaling decisions with the future workload forecasting results and the performance model. The scaling plan includes the number of pods and times that should be added or reduced.

\begin{figure*}[htbp]
\centering  
\subfigure[Actual $\#$ pods with pending]{\label{WP}
\includegraphics[width=0.35\textwidth]{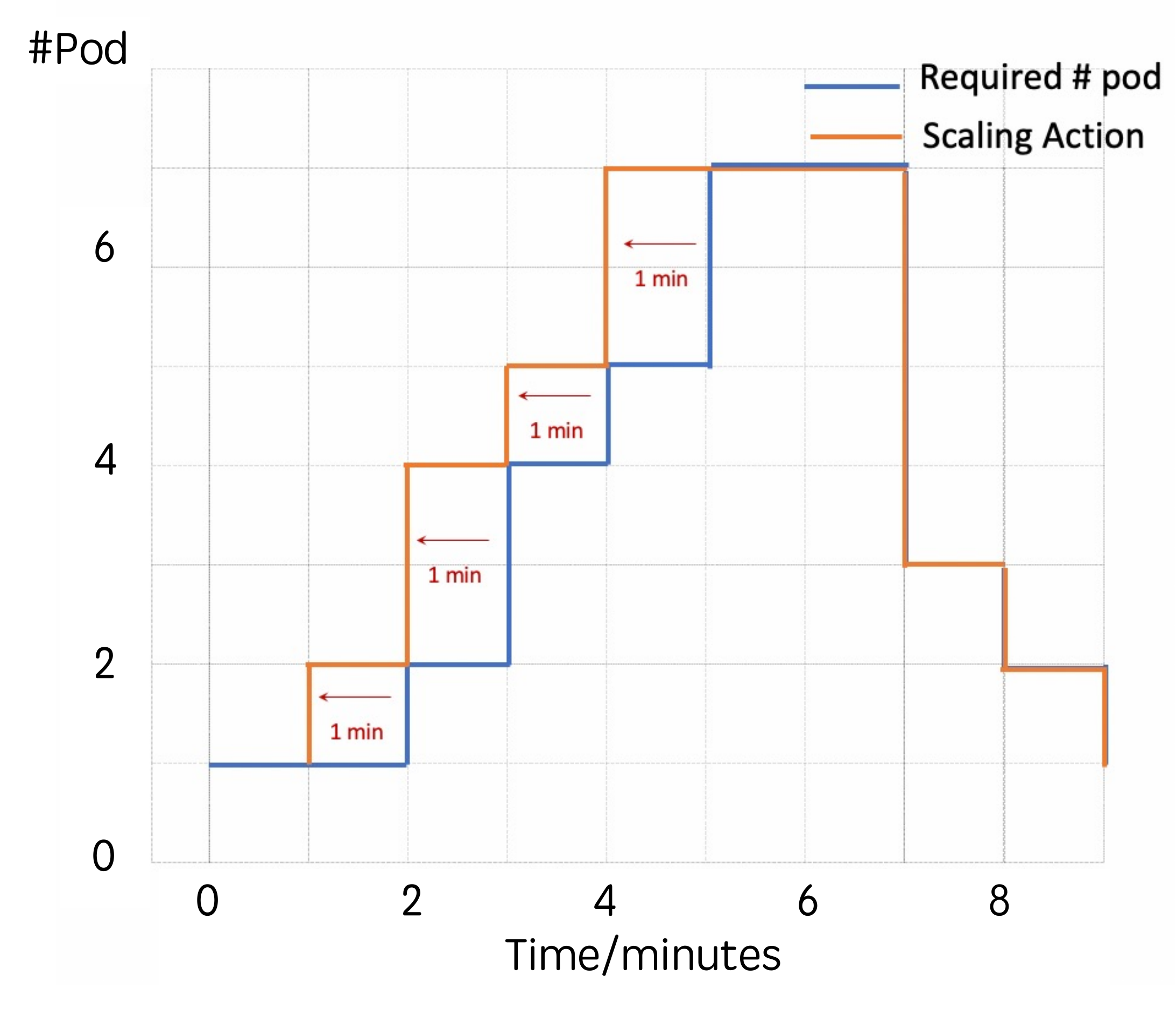}}
\subfigure[Pods scaling action]{\label{WOP}
\includegraphics[width=0.35\textwidth]{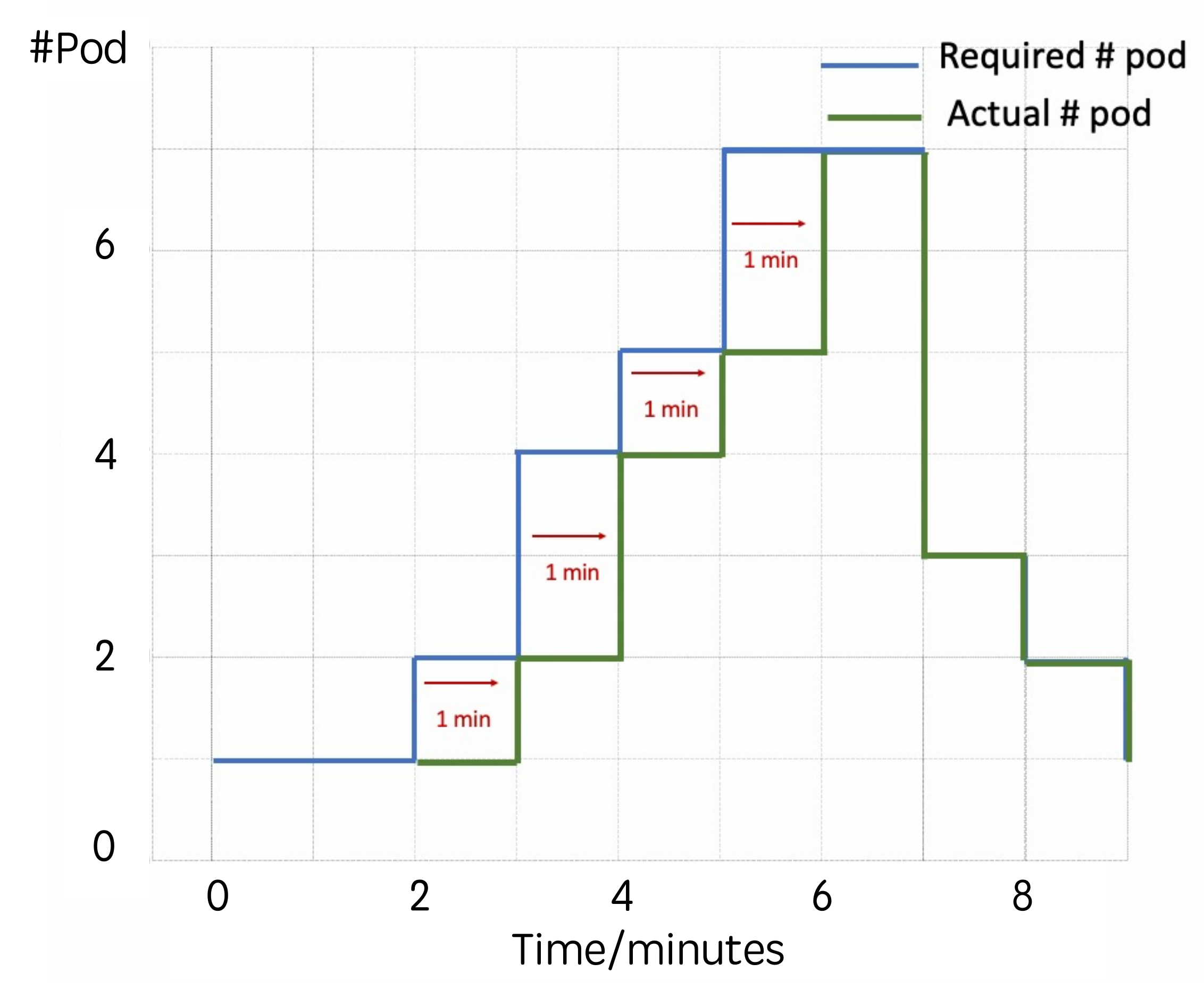}}
\caption{An example of the scaling action plan.}
\label{pending}
\end{figure*}

First, we forecast the number of pods required to satisfy clients' requirements in certain metrics (e.g., RT or CPU usage rate). However, there is an overhead time for pods to start, which means the system is troubled with time delay problems when adding resources, and the deployment throughput is limited. Facing the constraints mentioned above, we take the improved forecasting shift algorithm~\cite{flunkert2020simple} to handle such a challenge. 

A simple example is shown in Fig.~\ref{pending}. Assume the pending time (including pods starting and so on) is 1 minute; Fig.~\ref{WP} shows that if the needed pods are scaling precisely in real-time as required, the actual number of pods available will be delayed. Therefore, the pod scaling actions should be done in advance, as Fig.~\ref{WOP} shows. 

However, for the sake of stability, frequent actions are not a good choice. Generally, the operating frequency limit is set as one per every 3 minutes or every 5 minutes maximum. Thus, how to combine the scheduling plans within the corresponding interval is also a question worth exploring.

%% file: 4_experimental_evaluation.tex
\section{Experimental Evaluation}
Proper validation of the AHPA system is challenging, as these models are designed to adjust the horizontal pod schedule for different deployments. It is not possible to apply different models to one application at the same time. We here provide the real historical data collected from several deployments in Alibaba Cloud Container Service for Kubernetes (ACK) version 1.20.11 with Aliyun Linux/CentOS operating system. The time series data length is 20160, i.e., one data point per minute for 14 days.
The numerical results compared with the classical HPA algorithm and FixPod policy are shown in the following subsections.

\begin{figure}[t]
\centering 
\includegraphics[width=0.45\textwidth]{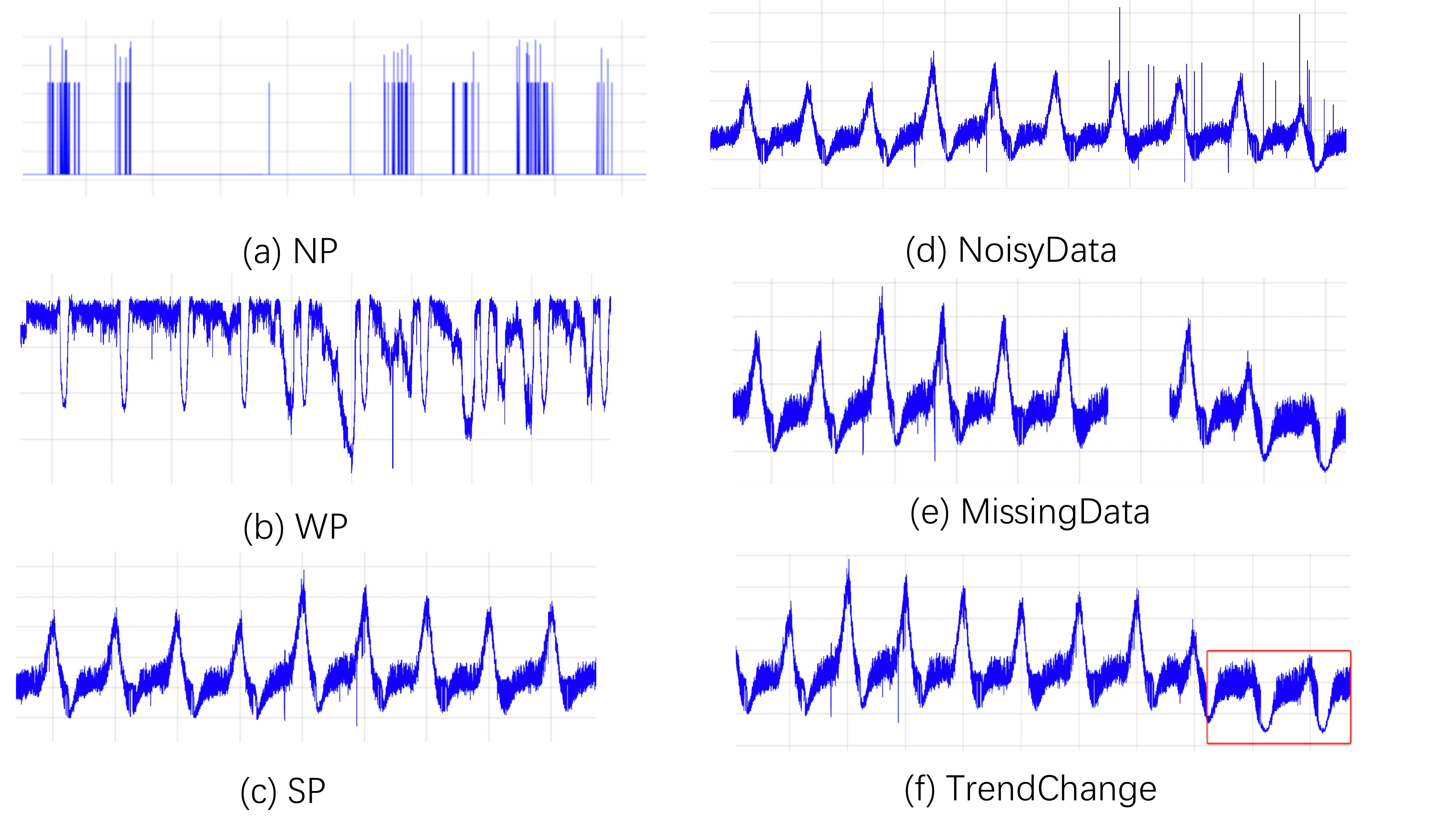} 
\caption{Typical signal series sampled from six actual datasets. } 
\label{ExpData} 
\end{figure}

As illustrated in Fig.~\ref{ExpData}, the dataset \textbf{NP}/\textbf{WP}/\textbf{SP} denote three scenarios that time series data with no periodicity, weak periodicity and strong periodicity, respectively. As for the \textbf{NoisyData} dataset, many outliers appear due to occasional network jitter caused by underlying network link reconstruction in the system. As mentioned above, missing values are common accidents due to some damage in nodes or monitor systems; we here test the framework by using the \textbf{MissingData} dataset, which drops all the data points on any random day. At last, we also consider another common challenging situation: the trend of time series changes, mostly caused by the new application version release. Hence, we would like to evaluate the performance of the AHPA framework when applied to the \textbf{TrendChange} dataset.

Since all models in a cloud container system aim to minimize the cost of satisfying the quality of service (QoS), we consider the following three metrics for evaluating these models: cost, violation rate (VR), and max pod number. The violation rate (\textbf{VR}) is used to evaluate the QoS of each model, i.e., for a specific metric target such as keeping CPU usage less than $50\%$, then VR would be computed as the time of CPU usage more significant than $50\%$ divided by the whole time length. The \textbf{cost} would be the total number of pods provided over time, computed as the integral of the number of pods over time (minute). According to the service-level agreement (SLA), the contract between a service provider and its customers that documents what services the provider will furnish and defines the service standards the provider is obligated to meet, the violation rate (\textbf{VR}) plays the most crucial role among all the criteria in the cloud service.

\subsection{Comparison Experiment}
In this subsection, we consider the performance of the AHPA framework under different situations of complex data, such as the time series data with strong periodicity, week periodicity, or without periodicity. 

\begin{figure*}[htbp]
\centering 
\includegraphics[width=0.8\textwidth,height=3cm]{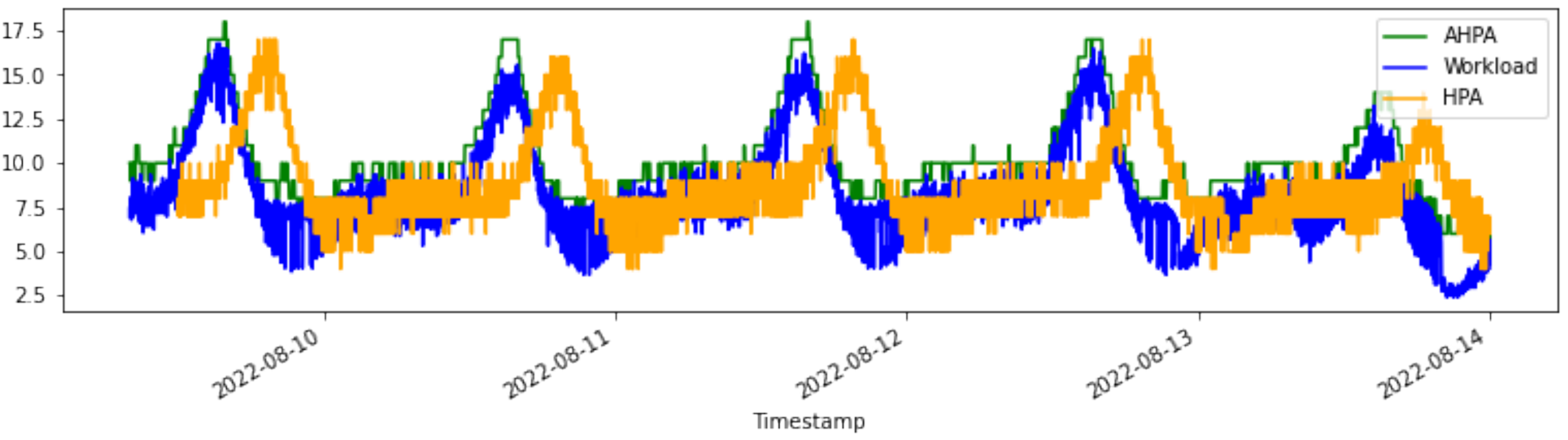} 
\caption{Comparison of the number of pods. } 
\label{Comparison} 
\end{figure*}

\begin{table*}[t]
\centering
\scalebox{0.85}{
 \begin{tabular}{c||c c c|c c c|c c c }
    \hline
        DataSet & \multicolumn{3}{|c|}{\textbf{NP}} &\multicolumn{3}{|c|}{\textbf{WP}}
       &\multicolumn{3}{|c}{\textbf{SP}} \\
       \hline
       Metric  & FixPod & HPA & AHPA & FixPod & HPA & AHPA & FixPod & HPA & AHPA  \\
       \hline
       Cost & 48830 & 19657& 22063 & 604680 & 447096 & 495541 & 201340 & 77464 & 92930\\
       \hline
       VR & 0 & 0.0047 & 0.0045 & 0 & 0.3975 & 0.0695 &0 & 0.2950 & 0.0266\\
       \hline
       Max Pod & 10 & 10 & 11 &60& 60 & 67& 20& 17& 18\\
       \hline
    \end{tabular}\label{table: periodicity test}
    }
\caption{Comparison of metric/cost/VR/max pod under different data structure.}
\label{CR_with_DDS}
\end{table*}
From the results shown in Table~\ref{CR_with_DDS}, we can observe that the FixPod policy provides a zero violation strategy but with a massive waste of pod resources in all cases. However, comparing the HPA and AHPA frameworks, the results demonstrate that AHPA achieves a much better QoS schedule plan than HPA with comparable costs. The advantage of AHPA is further highlighted when applied to time series with a more substantial periodical property.
In Fig.~\ref{Comparison}, the green and orange line represent the number of pods suggested from the AHPA and HPA method, and the blue line stands for the optimal number of pods when given the real workload and target metric value (e.g., keeping CPU usage under 50\%). We can obviously observe that adjustment from HPA is always lagging, which could lead to poor service experience at the peaks of workload. Moreover, from the comparison between the green and orange lines, we see the green line is much smoother than the orange line, which shows another advantage of the AHPA algorithm since the smoother of pod number is, the less cost spent on the operations including expanding and shrinking the pods resource.

\subsection{Robustness Experiment}
In this subsection, we evaluate the robustness of the AHPA framework due to the complex situations in real cloud services. Here we implement the algorithm on the actual data with noise, outliers, missing values, and abrupt trend changes.
\begin{table}[]
\centering
\scalebox{0.8}{
 \begin{tabular}{c|| c c| c c| c c }
    \hline
        DataSet & \multicolumn{2}{|c|}{\textbf{NoisyData}} &\multicolumn{2}{|c|}{\textbf{MissingData}}
       &\multicolumn{2}{|c}{\textbf{TrendChange}} \\
       \hline
       Metric  & HPA & AHPA  & HPA & AHPA  & HPA & AHPA  \\
       \hline
       Cost  & 77590& 93011  & 75966 & 92519 & 77464 & 92930\\
       \hline
       VR  & 0.2958 & 0.0276  & 0.2934 & 0.0307  & 0.2950 & 0.0266\\
       \hline
       Max Pod  & 31 & 28 & 17 & 18 & 17& 18\\
       \hline
    \end{tabular}
    }
\caption{Comparison of robustness test.}\label{table: robustness test}
\end{table}

From the results in Table~\ref{table: robustness test}, we can observe the violation rate of the HPA framework is around $30\%$, which means almost one-third of the time the QoS is worse than customers' expectation. It is almost $1000\%$ higher than the one of AHPA. Besides, from the results of the AHPA framework, we conclude the AHPA framework performs robustly in both violation rate and cost saving under noisy data, missing values, and trend-changing situations.

%% file: 5_application_use_payoff.tex
\section{Application Use and Payoff}

%

AHPA system has been deployed across Alibaba Cloud Services since April 2021 and has been promoted to many different cloud service scenarios from March 2022. Many related servers have used it to manage the elastic pod resources. Up to now, AHPA has been implemented in multiple customer scenarios, covering logistics, social networking, AI audio and video, e-commerce, online education, sports live+, and retail. Most such applications need to handle the challenges of real-time high-performance, low latency, and large and periodic business load fluctuations. In this section, we will discuss the impact of the AHPA system.

Take two applications as an example. For online education, ACK with AHPA system provides services with the following features: minute-level deployment, industry-leading reliability (a commitment of 99.999\% availability for individual instances), easy management, and high scalability (scaling according to real-time demands). For live streaming of e-commerce, services satisfying properties of fast deployment at a low-cost, ultra-low latency (2000+ nodes with a bandwidth of 150Tbps, bringing streaming latency less than 2 seconds) are provided by our system.

After customer business applications were deployed with the AHPA system, the elastic lag problem was considerably eased, with CPU usage increasing by 10\% and resource costs reduced by more than 20\%. At present, our AHPA algorithm has obvious advantages over other models. In addition, it can automatically perform flexible planning according to the changing trend of business volume without manual intervention, which significantly saves operation and maintenance costs. It is not easy to use digital numerical calculations for the workforce regarding operation and maintenance costs, but an automated AHPA algorithm greatly liberates the workforce and makes the entire system more efficient and universal. Specifically, for example, in the scenario of intelligent voice, AHPA is currently called around six thousand times a day and performs about 1,000 dispatches. Compared with another original management method, AHPA's result has reached that CPU usage is increased by about 9\% and saves the daily cost of POD resources (pieces * minutes) about ten thousand. With the help of AHPA, the AI voice business saves around 28\% of the original cost. 

%% file: 6_application_development_deployment.tex
\section{Application Development and Deployment}

\begin{figure*}[htbp]
\centering 
\includegraphics[width=0.68\textwidth]{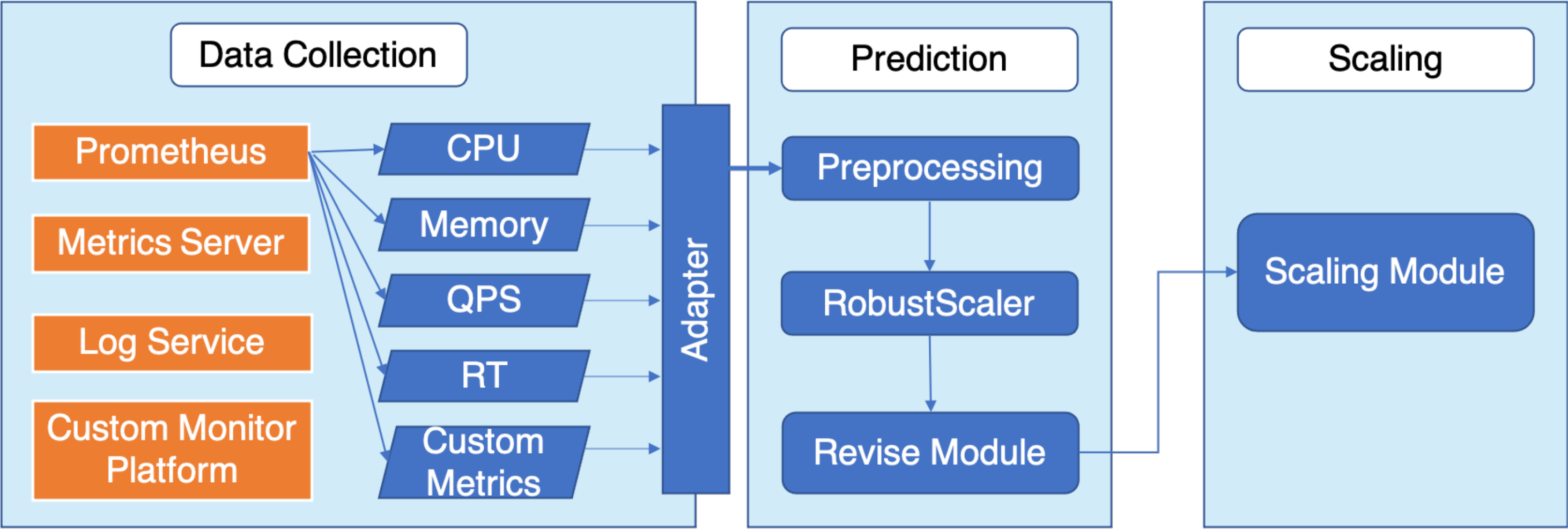} 
\caption{The framework of AHPA.} 
\label{Framework} 
\end{figure*}

\begin{figure*}[htbp]
\centering 
\includegraphics[width=0.68\textwidth]{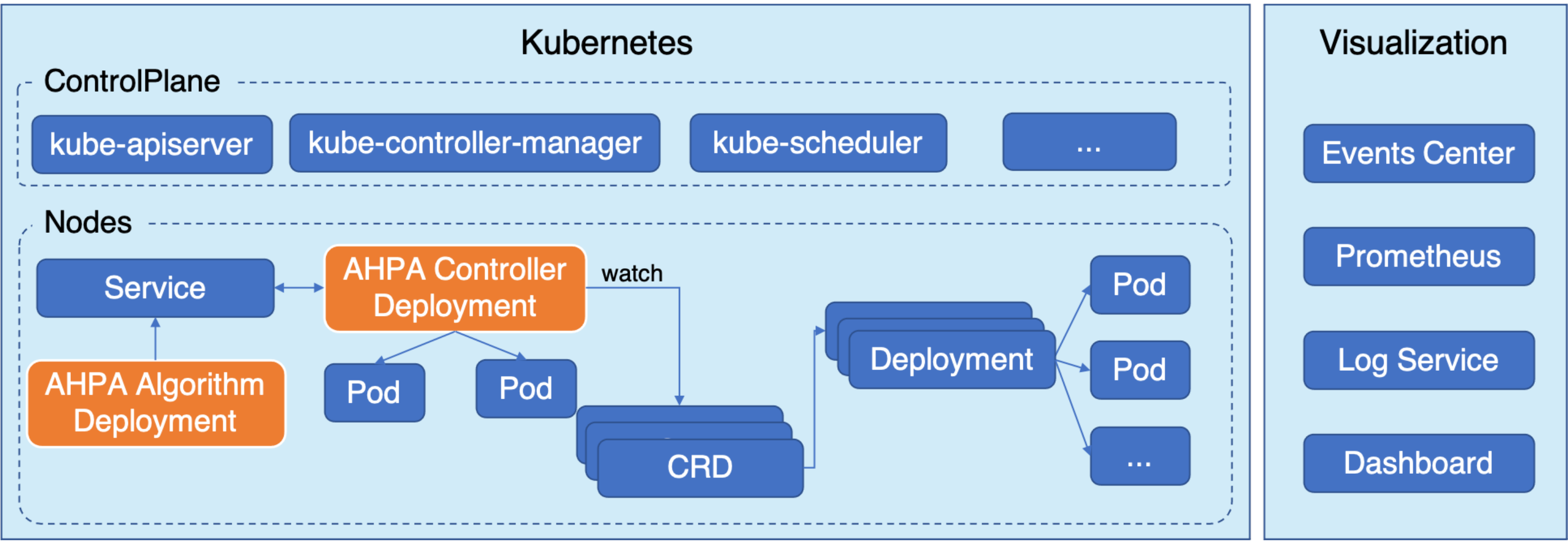} 
\caption{The deployed architecture of AHPA in Kubernetes.} 
\label{Deploy} 
\end{figure*}

Section \emph{Use of AI Technology} introduces the details of AI technology used in the algorithm part of the AHPA system. This section will show how such an algorithm is developed and deployed in the real cloud container system. We will first review the main components of AHPA (Fig.~\ref{Framework}) and then show the details of how it works in the whole Kubernetes system (Fig.~\ref{Deploy}).

\subsection{System Framework}
As shown in Fig.~\ref{Framework}, the framework of AHPA contains three main parts: Data Collection, Prediction, and Scaling. The main technologies and details of the Prediction part are introduced in section~\emph{Use of AI Technology}. More specifically, Preprocessing is the Data Preprocessing module in Fig.~\ref{Alg}, the scaling part contains Workload Forecasting and Performance Model Training module in Fig.~\ref{Alg}, and the Revise Module consists of the Scale Plan Generation module in Fig.~\ref{Alg}. 

Before Prediction, the Data Collection module collects data from various sources and transforms data into a unified form. Collected data consist of resource metrics (CPU, Memory, etc.), custom metrics (QPS, RT, etc.), and other external metrics. The data sources include Prometheus (an open-source system monitoring and alerting toolkit which stores metrics information as time series data), Metrics Server (a scalable, efficient source of container resource metrics for K8s), Log Service (a complete real-time data logging service developed by Alibaba Group) and other custom monitor platforms. After that, the Adapter module transforms different metrics into a unified form. 

The final module of AHPA is Scaling, which is used to scale pods according to the resource estimation results. Two scaling mechanisms are provided: Auto and Observer. Auto scales the number of pods according to the estimation results; the observer is the dry-run mode provided to the clients to monitor whether AHPA works as expected. 


\begin{figure}[htbp]
\centering 
\includegraphics[width=0.46\textwidth]{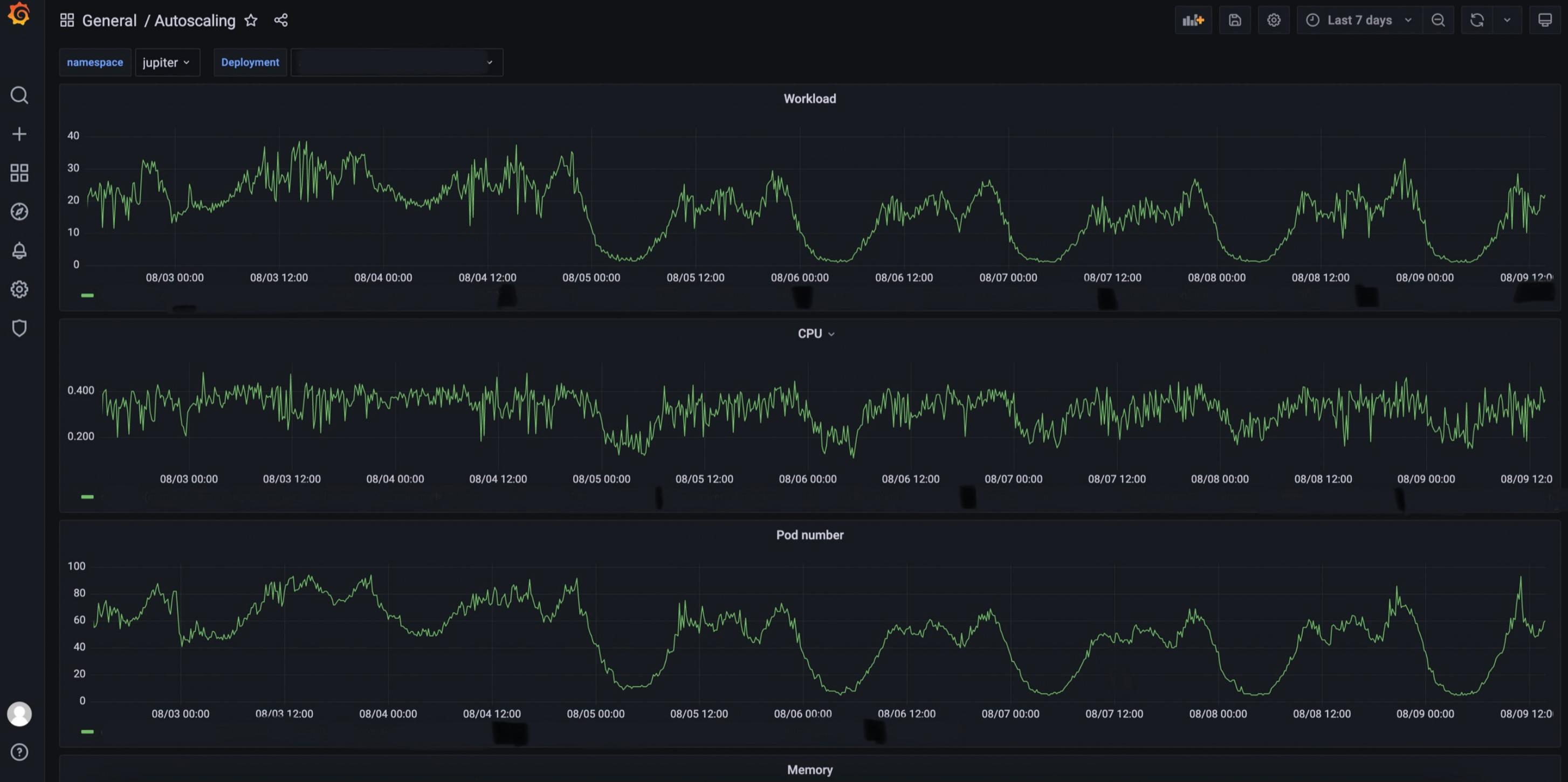} 
\caption{An example of visualization in the dashboard of AHPA.} 
\label{example of system-1} 
\end{figure}




\subsection{System Deployment}
AHPA is deployed on Alibaba Cloud Container Service for Kubernetes (ACK) with Go Programming Language. The deployment architecture of AHPA is shown in Fig.~\ref{Deploy}. There are two main components: AHPA Algorithm Deployment and AHPA Controller Deployment. 

CustomResourceDefinition (CRD) is introduced to deploy the pod scaling. CRD is flexible for different configures. The core parameters of CRD are following:
\begin{itemize}
    \item scaleTargetRef: positional, deployment for special objects. 
    \item metrics: positional, scaling metrics, e.g., CPU, Memory, RT, QPS, GPU, etc.
    \item averageUtilization: positional, the threshold of target. For instance, 40 means the utilization rate of the CPU can not be larger than 40\%.
    \item scaleStrategy: optional, scaling mechanism, Auto or Observer. Auto means pod scaling is deployed and Observer means not deploying pod scaling and only observing whether AHPA works as expected.
    \item maxReplicas: positional, the maximum number of instances to do scaling. 
    \item minReplicas: positional, the minimum number of instances to do scaling.
    \item instanceBounds: optional, time bounds of scaling, including start time and end time.
    \item cron: optional, setting for timing task.
\end{itemize}

Unique designs are also applied in deployment for \textbf{high availability} of AHPA service. Although anomalies always happen in complex systems. our goal is to provide high availability. When failures happen in pods, the failed pod will be killed, and a new pod will be created. 
Moreover, with the number of applications or services increasing, both the algorithm deployment part and controller deployment part can scale horizontally to satisfy the requirement of high concurrency. 

To ensure the business task is not aware of the update of AHPA, Algorithm and Controller communicate through service, and they can update independently. The update is rolling through pods, meaning the old pods will be killed only when the newly created pods work well as expected. 

\begin{figure}[htbp]
\centering 
\includegraphics[width=0.46\textwidth]{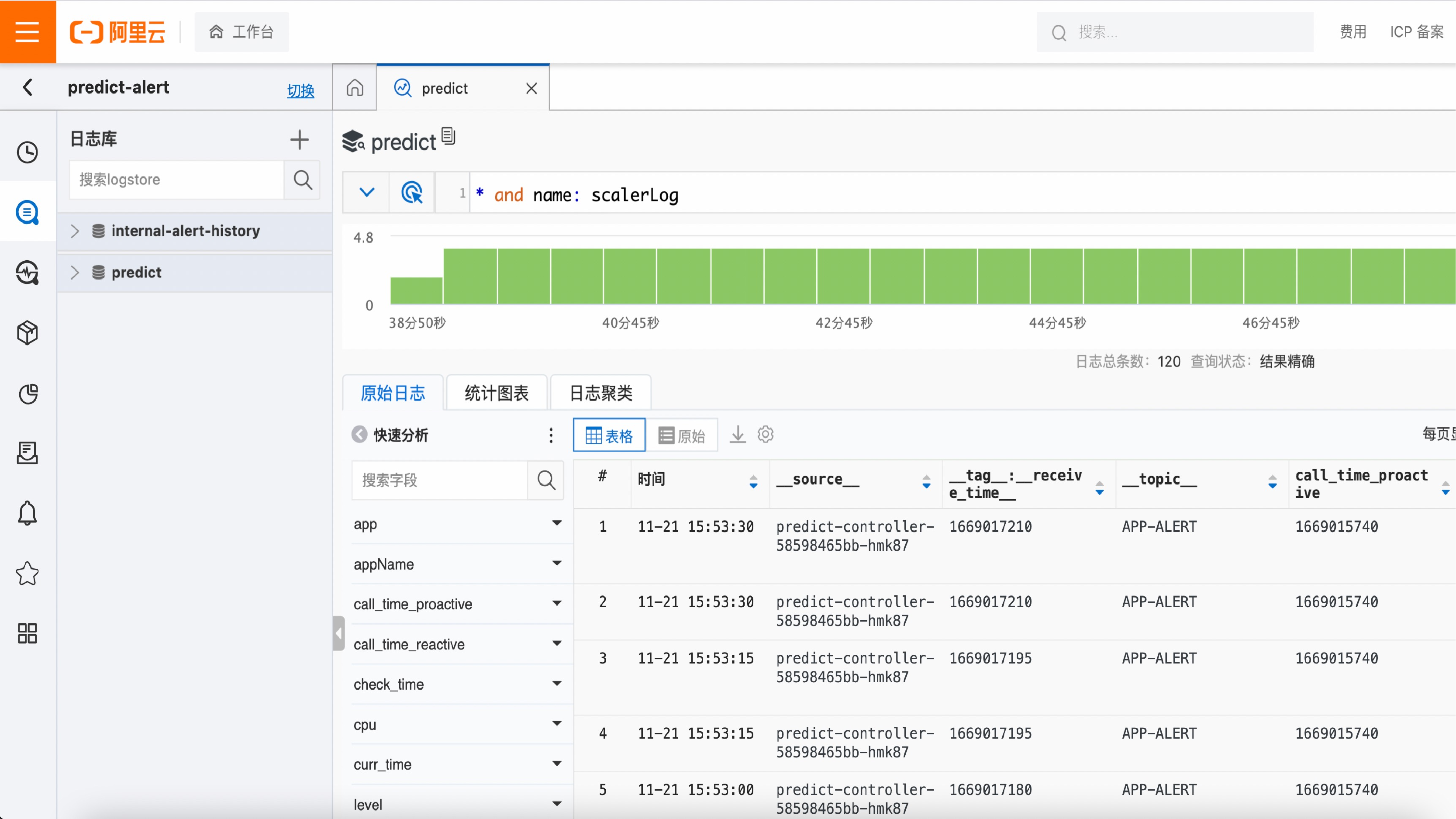}
\caption{An example of visualization in log service for AHPA.} 
\label{example of scheme} 
\end{figure}

Abundant \textbf{visualization} components are provided to help clients monitor the state of AHPA, including Kubernetes Event Center, Prothemetheus, Log Service, Dashboard, etc. Visualization examples of Dashboard and Log Service are shown in Fig.~\ref{example of system-1} and Fig.~\ref{example of scheme}, respectively. 

From the panel interface of Dashboard (shown in Fig.~\ref{example of system-1}), critical metrics, including Workload, CPU, memory, can be monitored in real-time. In addition, log Service (shown in Fig.~\ref{example of scheme}) provides information on every scaling action, including the time of scaling action, the mode (dry-run or not) of scaling action, the number of actual pods, pods number limitations, etc. All of these contribute to helping clients to monitor if AHPA works as expected and locate the cause of unexpected failures timely.

%% file: 7_maintenance.tex
\section{Maintenance}

As time goes by, changes in the data flow characteristics require dynamic tuning of the hyper-parameters of our models. However, as a whole, AHPA does not require a lot of post-operation and maintenance modifications. That is, in the workload forecasting module, due to the self-learning ability of deep learning models, after adding new data, there is no need to make many modifications to the original model, but it can be fine-tuned or re-training in response to the new data. Furthermore, in the following two modules, the final pod number adjustment scheme can also be adjusted according to the actual situation and does not involve model modification. The core idea of the entire algorithm is to predict the business as accurately as possible and deal with fluctuations in advance to save resources. Therefore, after AHPA is launched, it is unnecessary to do any significant modifications. In our practice,  only some model parameters need to be fine-tuned by April 2021.

%% file: 8_conclusion.tex
\section{Conclusions and Future Work}

In this paper, we present our improved framework AHPA (Adaptive Horizontal Pod Auto-scaling) for better resource management in Alibaba Cloud Container Service for Kubernetes to save resources while maintaining user experiences. This platform has been widely spread throughout Alibaba Cloud Services and deployed in various business scenarios not restricted to ACK container controlling. By accurately predicting the business volume of the next moment and obtaining the mapping relationship from historical data, AHPA could address the challenge of saving resources while ensuring business stability. The core section of the AHPA algorithm uses the robust decomposition-based time series forecasting module and the queue theory for performance modeling. Since its deployment in April 2021, AHPA has helped over 20 different business scenarios in Alibaba Cloud Services to solve the elastic lag problem and significantly increased the efficiency of CPU by 10\%. Besides, AHPA has demonstrated significant advantages in that it could automatically give the system's planning without any manual intervention. Further plans to expand AHPA in other parts of Alibaba Cloud have been scheduled. 

In subsequent work to adapt to different business scenarios, 
we will investigate how to abstract the mapping relationships in appropriate theories other than queuing theory according to the detailed requirements. Furthermore, with the vigorous development of cloud services, more and more large demands are constantly emerging. How to provide a framework with a higher degree of adaptation while ensuring the efficiency of each specific business scenario is also our pursuing goal. 

